\documentclass[lettersize,journal]{IEEEtran}
\usepackage{amsmath,amsfonts}
\usepackage{algorithmic}
\usepackage{algorithm}
\usepackage{array}
\usepackage[caption=false,font=normalsize,labelfont=sf,textfont=sf]{subfig}
\usepackage{textcomp}
\usepackage{stfloats}
\usepackage{url}
\usepackage{verbatim}
\usepackage{graphicx}
\usepackage{cite}
\usepackage{xcolor}
\usepackage{amssymb}
\usepackage{mathtools}
\usepackage{threeparttable}
\usepackage{bm}
\usepackage{multirow}
\usepackage{booktabs}
\usepackage{makecell}

\begin{document}

\title{D-PAD: Deep-Shallow Multi-Frequency Patterns Disentangling for Time Series Forecasting}

\author{Xiaobing Yuan, Ling Chen

\thanks{{Corresponding author: Ling Chen}.}
\thanks{Xiaobing Yuan and Ling Chen are with the State Key Laboratory of Blockchain and Data Security, Zhejiang University, Hangzhou 310027, China, and also with the College of Computer Science and Technology, Zhejiang University, Hangzhou 310027, China (e-mail: xybbo5@zju.edu.cn; lingchen@cs.zju.edu.cn).}}

\markboth{Journal of \LaTeX\ Class Files,~Vol.~14, No.~8, December~2023}%
{Shell \MakeLowercase{\textit{et al.}}: A Sample Article Using IEEEtran.cls for IEEE Journals}


\maketitle

\begin{abstract}
  In time series forecasting, effectively disentangling intricate temporal patterns is crucial. While recent works endeavor to combine decomposition techniques with deep learning, multiple frequencies may still be mixed in the decomposed components, e.g., trend and seasonal. Furthermore, frequency domain analysis methods, e.g., Fourier and wavelet transforms, have limitations in resolution in the time domain and adaptability. In this paper, we propose \textbf{D-PAD}, a \textbf{D}eep-shallow multi-frequency \textbf{PA}tterns \textbf{D}isentangling neural network for time series forecasting. Specifically, a multi-component decomposing (MCD) block is introduced to decompose the series into components with different frequency ranges, corresponding to the “shallow” aspect. A decomposition-reconstruction-decomposition (D-R-D) module is proposed to progressively extract the information of frequencies mixed in the components, corresponding to the “deep” aspect. After that, an interaction and fusion (IF) module is used to further analyze the components. Extensive experiments on seven real-world datasets demonstrate that D-PAD achieves the state-of-the-art performance, outperforming the best baseline by an average of 9.48\% and 7.15\% in MSE and MAE, respectively.
\end{abstract}

\begin{IEEEkeywords}
 Time series forecasting, disentanglement, decomposition and reconstruction.
\end{IEEEkeywords}

\section{Introduction} \label{sec:intro}

\IEEEPARstart{T}{ime} series forecasting is crucial in many real-world applications, e.g., traffic~\cite{chen2023multi}, finance~\cite{lai2018modeling}, and energy~\cite{li2019enhancing}. In real-world time series, multiple patterns, including trend, seasonality, and other hidden patterns, are entangled and make forecasting a challenging task.

To address this challenge, many researchers have devoted their efforts to decomposing time series into few components, each representing an underlying pattern~\cite{hyndman2018forecasting,wu2021autoformer}. Traditional methods~\cite{hyndman2018forecasting, gardner1985exponential, taylor2018forecasting} decompose time series into trend and seasonal components and make predictions based on specific reasoning rules. With the development of deep learning, some researchers combine traditional methods with deep models~\cite{zeng2023transformers,zhou2022fedformer,cui2016multi}, which input the trend and seasonal components obtained from decomposition into neural networks. While some researchers directly endow deep models themselves the ability to disentangle by progressive decomposition~\cite{wu2021autoformer}, supervision by contrastive learning~\cite{woo2022cost}, variational inference~\cite{wang2022learning}, etc., which can capture temporal patterns in more flexible representations.

Despite the success, these methods only focus on trend and seasonal components in time series, ignoring other hidden patterns, which results in failing to disentangle the patterns of multiple frequencies. To this end, some researchers turn to the frequency domain~\cite{zhou2022fedformer,woo2022cost, yang2020adaptive,minhao2021t,deznabi2023multiwave}, whereas, they suffer from poor resolution in the time domain and obvious lack of adaptability. Others employ deep stacked models consisting of fully connected layers~\cite{oreshkin2019n,challu2023nhits,fandepts} to extract multiple components in a hierarchical order. However, in the end-to-end architectures constrained only by residual connections, much information of the same frequencies may be scattered and left in different components.

Empirical Mode Decomposition (EMD)~\cite{huang1998empirical} is widely used in signal analysis, image processing, speech recognition, and other fields due to its adaptability, directness, and intuitiveness. EMD decomposes a signal into intrinsic mode functions (IMFs) in multiple frequency ranges. However, exiting methods~\cite{jiang2018time,huang2014monthly,kim2018identification, TowardsDiverse} only use EMD for preprocessing in time series analysis and modeling, as it is not naturally formulated in the neural network paradigm~\cite{velasco2022learnable}. In addition, the iterative sifting process of EMD may cause the mixing of patterns of different frequencies into the same IMF.

To address the aforementioned problems, we propose \textbf{D-PAD}, a \textbf{D}eep-shallow multi-frequency \textbf{PA}tterns  \textbf{D}isentangling neural network for time series forecasting. To the best of our knowledge, D-PAD is the first work that explicitly captures the temporal patterns of multiple frequency ranges from multiple components, and learns the information of the same frequencies scattered and mixed in various components via the “shallow” and “deep” disentanglement of temporal patterns. The major contributions of this work are outlined as follows:
\begin{itemize}
    \item Introduce a multi-component decomposing (MCD) block to achieve the “shallow” disentanglement of intricate temporal patterns, which breaks the convention of using EMD as a data preprocessing step with the morphological operators, and provides an adaptive and progressive approach to capture the temporal patterns of multiple frequency ranges with high resolution in the time domain.
    
    \item Propose a decomposition-reconstruction-decomposition (D-R-D) module to achieve the “deep” disentanglement of temporal patterns, which self-separates and reconstructs the components obtained from “shallow” disentanglement, and further decomposes the reconstructed sequences in following MCD blocks, thereby learning the information of the same frequencies scattered and mixed in various components.
    
    \item Conduct extensive experiments on seven real-world time series datasets. The results show that D-PAD outperforms the best baseline by an average of 9.48\% and 7.15\% in MSE and MAE, respectively.
\end{itemize}

\section{Related Work} \label{sec:related}

\subsection{Deep Time Series Forecasting}

Time series forecasting has been extensively studied in the past decades, and deep models have shown promising results, e.g., multi-layer perceptrons (MLPs), recurrent neural networks (RNNs), and temporal convolution networks (TCNs). MLP-based models~\cite{oreshkin2019n, challu2023nhits, fandepts, zeng2023transformers} encode the temporal patterns into the fixed parameter of MLP layers along a specific dimension. RNNs and their variants~\cite{chen2022multiscale} model the temporal patterns and predict iteratively, which have achieved great success. The works based on TCN~\cite{oord2016wavenet, liu2022scinet} introduce dilated causal convolutions to expand the receptive field, and model temporal patterns of different scales. For example, SCINet~\cite{liu2022scinet} utilizes a recursive downsample-convolve-interact architecture, which uses multiple convolutional filters to extract distinct yet valuable temporal features from the downsampled sub-sequences and features. Recently, Transformer-based models~\cite{chen2021learning, wu2021autoformer, zhou2021informer, zhou2022fedformer, li2019enhancing} have dominated this landscape, which take advantage of attention mechanism to discover the relationships across the sequence and focus on the important time steps. For example, LogTrans~\cite{li2019enhancing} introduces the local convolution to Transformer and utilizes the LogSparse attention to select time steps following the exponentially increasing intervals, which reduces the complexity. Informer~\cite{zhou2021informer} extends Transformer with KL-divergence based ProbSparse attention and greatly reduces the complexity. However, these deep models mainly focus on the original time series and do not learn disentangled representations of different frequency components, making it difficult to capture intricate temporal patterns effectively.

In our work, the proposed MCD blocks adaptively decompose the time series and their subsequences into multiple components in different frequency ranges to extract and model intricate temporal patterns.

\subsection{Decomposition of Time Series} 

Decomposition is an important way in time series analysis. Early methods, e.g., ARIMA~\cite{hyndman2018forecasting}  and ETS~\cite{gardner1985exponential}, mainly focus on decomposing time series into trend and seasonal components. The deep models follow this convention~\cite{wu2021autoformer, zeng2023transformers, wang2022learning}. For example, Autoformer~\cite{wu2021autoformer} makes series decomposition as basic inner blocks in the Transformer-based model to obtain trend and seasonal components progressively. DLinear~\cite{zeng2023transformers} combines decomposition with linear layers, which has a simple structure and achieves excellent performance. LaST~\cite{wang2022learning} uses variational inference to design the trend and seasonal representations learning and disentanglement mechanisms. However, due to their little or no consideration for other components, they suffer from the entangled patterns of multiple frequencies.

To address this problem, existing methods can be roughly divided into two categories based on the design philosophy. The first category~\cite{zhou2022fedformer,woo2022cost, deznabi2023multiwave, chen2022multiscale} is to model time series in the frequency domain, primarily using Fourier Transforms~\cite{bracewell1986fourier} and wavelet transforms~\cite{torrence1998practical}. For example, FEDformer~\cite{zhou2022fedformer} develops a frequency enhanced Transformer and achieves linear complexity by randomly selecting a fixed number of frequency components. Based on Fourier Transform, CoST~\cite{woo2022cost} comprises both time domain and frequency domain contrastive losses to learn discriminative trend and seasonal representations, respectively. MultiWave~\cite{deznabi2023multiwave} uses multi-level discrete wavelets to decompose each signal into subsignals of varying frequencies and groups them into different frequency bands. Nevertheless, these methods have significant limitations in terms of adaptability, and the effectiveness of modeling is also affected by information loss, the Gibbs effect, etc. The second category~\cite{oreshkin2019n,challu2023nhits,fandepts} is to design deep neural architectures based on residual connections and the deep stacks of fully-connected layers. For example, N-HITS~\cite{challu2023nhits} utilizes the doubly residual structure and basis expansion, which extracts and removes components layer by layer. Though these methods are not limited to focus on only a few components, they still fail to “deep” disentangle temporal patterns, as some information of different frequencies is still mixed in a component.

In our work, D-PAD reconstructs and re-decomposes the previous decomposed components by the D-R-D module, learning the information of the same frequencies scattered in different components for time series forecasting.

\subsection{EMD for Time Series Analysis} \label{sec:EMD_related}

EMD~\cite{huang1998empirical}, fully unsupervised, has been applied to various signal analysis tasks. EMD decomposes the original signal into IMFs by leveraging local characteristics, effectively handling both stationary and non-stationary signals. Compared with the frequency domain algorithms~\cite{bracewell1986fourier,torrence1998practical}, EMD can more accurately reflect the physical characteristics of the original signal and shows a stronger local performance. Therefore, in dealing with non-linear and non-stationary signals, EMD is more effective~\cite{jiang2018time,nunes2003texture}. Many researchers~\cite{jiang2018time,huang2014monthly,kim2018identification,TowardsDiverse} have already applied EMD for time series analysis. For example, M-EMDSVM~\cite{huang2014monthly} combines EMD and support vector machine and makes an improvement by removing the high frequency for monthly streamflow forecasting. STAug~\cite{TowardsDiverse} uses EMD, reassembles the subcomponents with random weights, and adapts a mix-up strategy that generates diverse as well as linearly in-between coherent samples. Although EMD serves as an effective preprocessing step for series decomposition, enabling the downstream models to analyze flexible representations, its two principal components, i.e., the detection of local extrema and the interpolation, are not naturally formulated in the neural network paradigm. This incongruity hinders the full potential of deep models to progressively decompose and effectively model temporal patterns~\cite{wu2021autoformer}. In addition, the interpolation always creates additional information that has nothing to do with the original data.

In our work, inspired by the development of EMD in image processing~\cite{bhuiyan2008fast,nunes2003image,el2009pde}, morphological operators are introduced to time series analysis. The aforementioned issues are addressed by using morphological EMD (MEMD) as an inner block of the deep model.

\section{Preliminaries}\label{preliminary}

\subsection{Problem Formulation}

Consider a time series with a lookback window of length $T$, denoted as $\bm{X}=\{x_{t-T+1},x_{t-T+2},\cdots,x_t\} \in \mathbb{R}^{T}$, where $x_{t}$ is the value at time step $t$. The objective is to forecast future $H$ values $\hat{\bm{X}}=\{\hat{x}_{t+1},\hat{x}_{t+2},\cdots,\hat{x}_{t+H}\} \in\mathbb{R}^{H}$. Therefore, the forecasting task can be formulated as follows:
\begin{equation}
    \hat{\bm{X}}=f(\bm{X},\mathbf{\Phi}),
\end{equation}
where $f$ is the deep learning network for the task. $\mathbf{\Phi}$ denotes all learnable parameters of $f$.

\begin{figure*}
\vspace{-5pt}
\begin{center}
	\centerline{\includegraphics[width=0.92\linewidth]{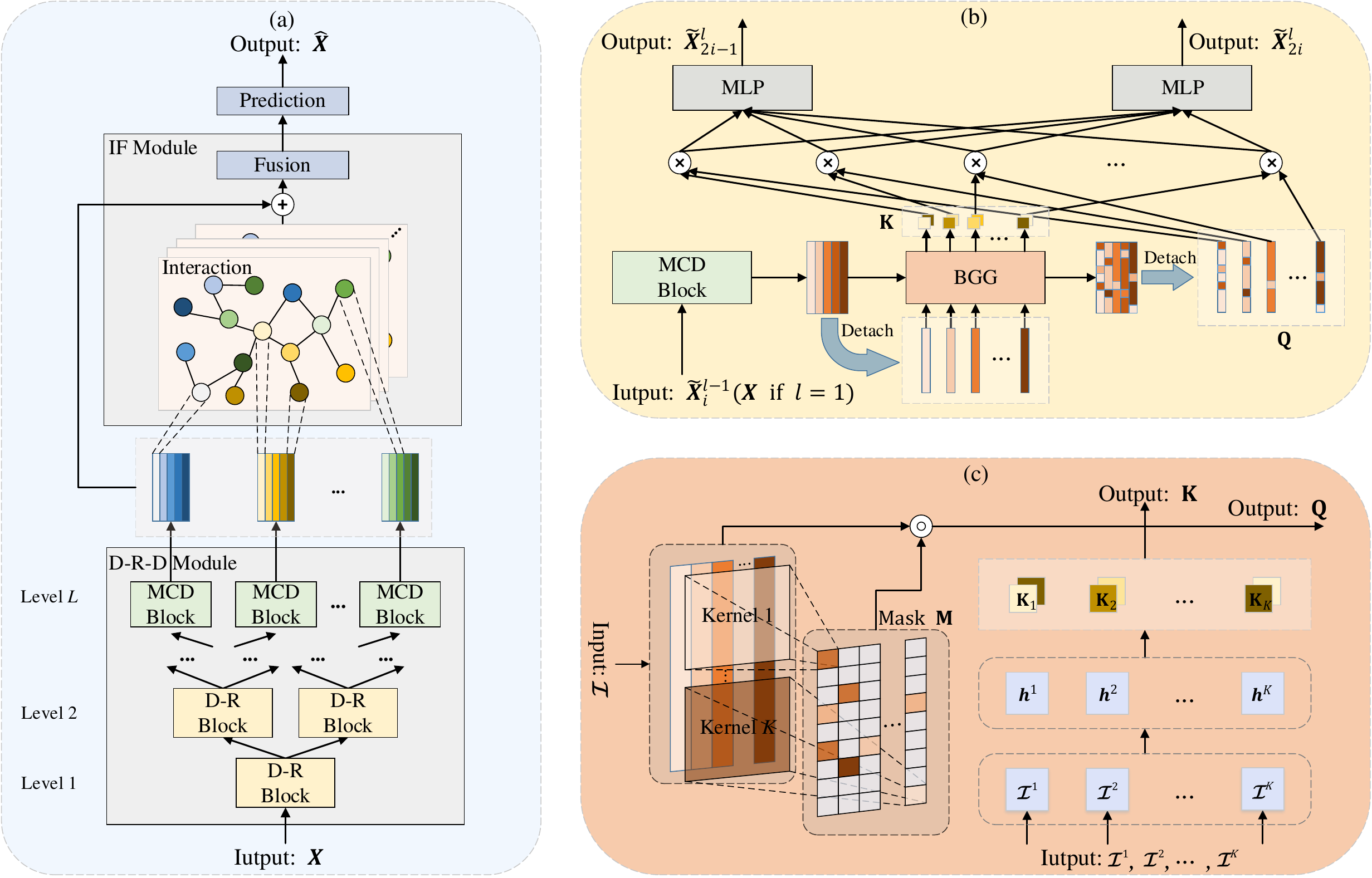}}
	\caption{Overview of D-PAD. (a) D-PAD is primarily composed of two parts, i.e., the D-R-D module and the interaction and fusion (IF) module. (b) The D-R block decomposes a series into multiple components and reconstructs them into two new series. (c) BGG is the combination of convolutions and projections, which generates $\mathbf{Q}$ and $\mathbf{K}$ to guide the branch selection for each component. (Best viewed in color).}
	\label{fig:framework}
\end{center}
\vspace{-5pt}
\end{figure*}

\subsection{Details of EMD}

{\bf{The overview of EMD:}}
As an adaptive method used for signal analyzing, EMD~\cite{huang1998empirical} decomposes an input signal into a finite sum of simpler signals (modes), named IMFs, i.e., given a signal $s(t)$, EMD is an iterative process as follows:
\begin{enumerate}
    \item[1)]Identify all the local extrema, including local maxima and local minima, of $s(t)$.
    \item[2)]Interpolate all the local maxima together to get the upper envelope $s_\mathrm{up}(t)$, and all the local minima together to get the lower envelope $s_\mathrm{low}(t)$. 
    \item[3)]Calculate the local mean as the average of both envelops: $m(t)=\frac{1}{2}\left(s_\mathrm{up}(t)+s_\mathrm{low}(t)\right)$.
    \item[4)]Extract the candidate IMF: $I^{\prime}(t)=s(t)-m(t)$.
    \item[5)]	Check the properties of $I^{\prime}(t)$: If $I^{\prime}(t)$ satisfies some characteristics, e.g., a selected tolerance criterion, an IMF $I(t)=I^{\prime}(t)$ is derived and meantime $s(t)$ is replaced with the residual $r(t)=s(t)-I(t)$; else $I^{\prime}(t)$ is not an IMF and $s(t)$ is replaced with $I^{\prime}(t)$.
    \item[6)]Repeat steps 1) - 5) until the residual satisfies the stop criterion.
\end{enumerate}

After finishing the process, the original signal can be expressed as follows:
\begin{equation}
    s(t) = {\sum_{i = 1}^{K}{I_{i}(t)}} + r(t),
\end{equation}
where $K$ is the number of IMFs, $I_{i}(t)$ represents the $i$-th IMF ordered by descending frequency, and $r(t)$ is the residual containing the central tendency information of the signal $s(t)$.

{\bf{The relative tolerance:}}
In the empirical mode decomposition process (EMP), the relative tolerance is a criterion for judging whether the candidate IMF is an IMF. It is a Cauchy-type stop criterion proposed in~\cite{wang2010intrinsic}, which is widely used in the implementation of EMD. The current relative tolerance is defined as follows:
\begin{equation}
    \mathrm{RT} = \frac{\|\bm{\mathcal{I}}^{\prime}_\mathrm{prev}-\bm{\mathcal{I}}^{\prime}_\mathrm{cur}\|^{2}_{2}}{\|\bm{\mathcal{I}}^{\prime}_\mathrm{prev}\|^{2}_{2}},
\end{equation}
where $\bm{\mathcal{I}}^{\prime}_\mathrm{prev}\in\mathbb{R}^{T}$ represents the previous candidate IMF and $\bm{\mathcal{I}}^{\prime}_\mathrm{cur}\in\mathbb{R}^{T}$ is the current candidate IMF. $\bm{\mathcal{I}}^{\prime}_\mathrm{cur}$ is considered as an IMF if $\mathrm{RT}\leq0.2$.

\section{Methodology}

\subsection{Overview}

Fig.~\ref{fig:framework} shows the overview of D-PAD. It aims to achieve a detailed but non-redundant disentanglement, i.e., decomposing time series into a finite number of representative components, each containing information within a similar frequency range. Specifically, the MCD block decomposes series into multiple components with different frequency ranges. The core of this block is MEMD, as shown in Fig.~\ref{fig:MCD}. To cope with information mixing, the reconstruction and branch selection of the components are guided by branch guidance generators (BGG) in decomposition-reconstruction (D-R) blocks, which are stacked to form the D-R-D module and enable a progressive decomposition of the time series. In addition, the interaction and fusion (IF) module incorporates interaction learning between the components obtained from the D-R-D module. Subsequently, the components are fused for prediction.

\subsection{MCD Block}

\begin{figure*}
\vspace{-8pt}
\begin{center}
	\centerline{\includegraphics[width=0.95\linewidth]{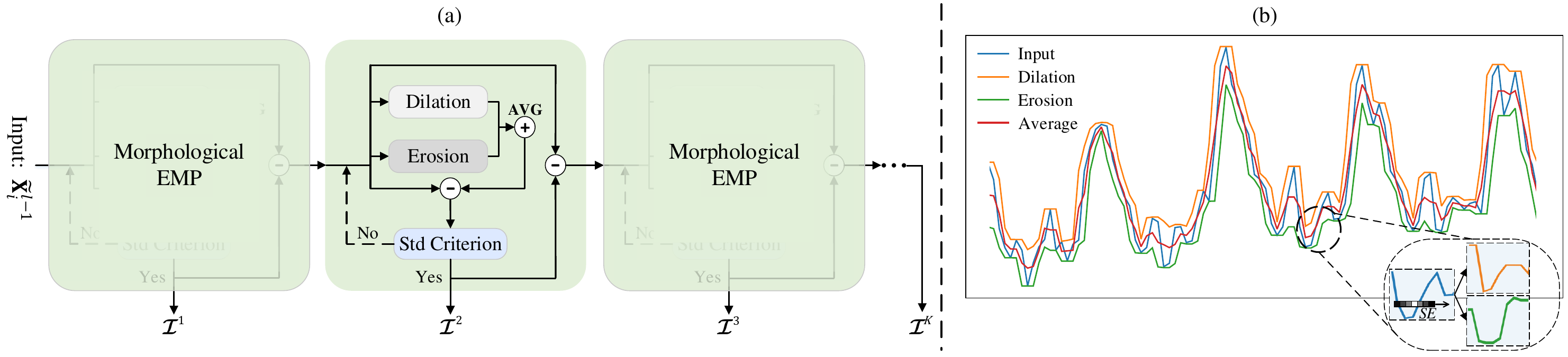}}
	\caption{MCD block and its diagram. (a) The core of the MCD block is MEMD, which includes the iterative morphological empirical mode decomposition process (EMP). (b) The mathematical morphology is employed to calculate and draw upper and lower envelope curves in MEMD. (Best viewed in color).}
	\label{fig:MCD}
\end{center}
\vspace{-15pt}
\end{figure*}

Fig.~\ref{fig:framework}(a) and~\ref{fig:framework}(b) show that the MCD block is the basic decomposition component of D-PAD, which achieves the “shallow” disentanglement of temporal patterns. In theory, it is necessary to consider the inductive bias of the method and dataset for disentangled representation learning~\cite{locatello2019challenging}. This can be reflected either inherently in the decomposition process, e.g., EMD, or explicitly in supervision within deep learning. As discussed in Section~\ref{sec:related}, existing deep learning methods often result in the mixing of information. Therefore, EMD naturally becomes an economical and effective choice for building MCD. However, the issue of extrema and interpolation, discussed in Section~\ref{sec:EMD_related}, hinders its potential in conjunction with neural network models. To tackle this dilemma, we utilize morphological operators in mathematical morphology~\cite{serra1982image}, i.e., dilation and erosion, to calculate and draw the upper and lower envelope curves of time series, as depicted in Fig.~\ref{fig:MCD}(b). This process is called MEMD and allows the MCD block to be integrated into neural networks and stacked in multiple layers. In order to adapt mathematical morphological operators to the field of time series, we give their definitions first.

\subsubsection{Dilation and erosion}
 Considering the extended-real-valued functions $s:\mathbb{E}\rightarrow\bar{\mathbb{R}}$, where $\bar{\mathbb{R}} = \mathbb{R} \cup \{ - \infty, + \infty \}$, we denote the set of all such functions as $\mathcal{S}(\mathbb{E},\bar{\mathbb{R}})$. The fundamental morphological operators, i.e., dilation and erosion, are actually a special case of the convolution in the max-plus algebra and its dual, respectively. Specifically, dilation enlarges the domain of the function, while erosion shrinks it.

{\bf{Definition 1:}}
The dilation $\delta_{SE}(s)$ of $s$, which is the same as the sup-convolution in convex analysis, is defined as follows: 
\begin{equation}
\begin{split}
    \delta_{SE}(s)(t) :&= \sup\limits_{t^\prime \in \mathbb{E}}\{ {s( t^\prime ) + SE( {t - t^\prime} )} \}  \\
    &= \sup\limits_{w \in \mathbb{E}}\{ {s( {t - w} ) + SE(w)} \}, \\
\end{split}
\end{equation}
where $SE\in S(\mathbb{E},\bar{\mathbb{R}})$ is the additive structuring function and the inf-addition rule $\infty-\infty=\infty$ is to be used in case of conflicting infinities. In mathematical morphology, the basic transformations of data are performed iteratively using basic symmetric structuring elements (SEs). The $\sup s$ and $\inf s$ refer to the supremum and infimum of $s$, respectively.

{\bf{Definition 2:}}
The erosion $\varepsilon_{SE}(s)$ of $s$, which is the same as the inf-convolution in convex analysis, is defined as follows:
\begin{equation}
\begin{split}
    \varepsilon_{SE}(s)(t):&=- \delta_{\check{SE}}( {- s} )(t) \\
    &= \inf\limits_{t^\prime \in \mathbb{E}}\{ s( t^\prime ) - SE( {t^\prime - t} ) \} \\
    &= \inf\limits_{w \in \mathbb{E}}\{ s( {t + w} ) - SE(w) \}, \\
\end{split}
\end{equation}
where $\check{SE}(t)=SE(-t)$ is the transposed structuring function. With the two aforementioned definitions, we can obtain the following two corollaries:

{\bf{Corollary 1:}}
In the field of time series analysis, where the samples are discrete, the $\sup$ and $\inf$ can be replaced by $\max$ and $\min$, respectively. The dilation $\zeta_{\acute{SE}}(x)$ of time series can be formulated as follows:
\begin{equation}
\begin{split}
    \zeta_{\acute{SE}}(x_{t}):&=\max\limits_{t^{\prime} \in [t - C,t + C]}( x_{t^{\prime}} + \acute{SE}_{t^{\prime} - t}) \\
    &= \max\limits_{\omega \in [- C,C]}( x_{t + \omega} + \acute{SE}_{\omega}), \\
\end{split}
\end{equation}
where $\acute{SE}\in\mathbb{R}^{2C+1}$ is an SE kernel with length $2C+1$, which is a substitute for the structuring function.

{\bf{Corollary 2:}}
Similarly, the erosion $\varsigma_{\acute{SE}}(x)$ of time series can be formulated as follows:
\begin{equation}
\begin{split}
    \varsigma_{\acute{SE}}(x_{t}) :&= - \zeta_{\acute{SE}}( {- x_{t}} ) \\
    &= {\min\limits_{t^{\prime} \in [t - C,t + C]}( {x_{t^{\prime}} - {\acute{SE}}_{t^{\prime} - t}} )} \\
    &= \min\limits_{\omega \in [- C,C]}( {x_{t + \omega} - {\acute{SE}}_{\omega}}), \\
\end{split}
\end{equation}
where the transposed SE kernel is equal to $\acute{SE}$, as the SE kernels are 1D and symmetric in our work. The dilation and erosion can be seen as the maximum and minimum filters of time series, respectively, which correspond well to the envelopes. Then we can define a morphological mean envelope as follows:
\begin{equation}
    \bm{m}_{\acute{SE}} := \frac{\zeta_{\acute{SE}}\left( \bm{X} \right) + \varsigma_{\acute{SE}}\left( \bm{X} \right)}{2},
\end{equation}
where $\bm{X}$ and $\bm{m}_{\acute{SE}} \in \mathbb{R}^T$ denote the input series and the average of the upper and lower envelopes, respectively. Then, the candidate IMF can be extracted by: $\bm{\mathcal{I}}^{\prime}=\bm{X}-\bm{m}_{\acute{SE}}$, which is considered as the $i$-th IMF: $\bm{\mathcal{I}}^{i}=\bm{\mathcal{I}}^{\prime}$ if it satisfies the relative tolerance criterion. Whenever an IMF is obtained, we consider one morphological EMP completed, as illustrated in Fig.~\ref{fig:MCD}(a).

\subsubsection{SE kernels}
It is evident that the structuring functions determine the effect of morphological operations, which can be considered as functions defined on a certain region and describe specific shape feature within that region. Due to its simplicity and efficiency, we choose the naïve SE: the zero SE, and represent the structuring functions in the form of SE kernels to implement MEMD on time series. The zero SE is often used to control the shape of morphological operations without changing the value of the function. The zero SE kernel $\bm{o}\in\mathbb{R}^{2C+1}$ is defined as follows:
\begin{equation}
    \bm{o}_{t^{\prime} - t} = 0,~t^{\prime} \in [t - C,t + C]~~or~~\bm{o}_{\omega} = 0,\left\| \omega \right\| \leq C.
\end{equation}

Then the morphological mean envelope is calculated as follows:
\begin{equation}
    \bm{m}_{\bm{o}_{\omega}} := \frac{\zeta_{\bm{o}_{\omega}}\left( \bm{X} \right) + \varsigma_{\bm{o}_{\omega}}\left( \bm{X} \right)}{2}.
\end{equation}

It avoids using the interpolation method and preserves the intrinsic characteristics of the input time series without generating additional information. Moreover, it greatly improves the efficiency of EMP by the convolution-like operator \textit{unfold ()} in PyTorch. Each IMF naturally corresponds to a component of time series. After completing all the morphological EMPs, the input series $\bm{X}$ is decomposed into $K$ components $\bm{\mathcal{I}}\in\mathbb{R}^{T\times K}$, where $\bm{\mathcal{I}}^{i}\in\mathbb{R}^{T}$ is the $i$-th component.

\subsection{D-R-D Module}
\label{sec:D-R-D}

With the MCD block presented above, we construct a multi-level decomposition and reconstruction module, i.e., the D-R-D module, which achieves the “deep” disentanglement of temporal patterns. It consists of the D-R blocks arranged in a tree structure, where BGG provides guidance for the selection of components obtained from the MCD block. There are $2^{l-1}$ D-R blocks at the $l$-th level, where $l=1,\cdots,L$ is the index of the level, and $L$ is the total number of levels.

\subsubsection{BGG}

Fig.~\ref{fig:framework}(c) shows the details of BGG, which consists of two parts: intra-projection for generating the key and inter-mask for generating the query. They focus on modeling the time dependency within a single component and among multiple components, respectively, and dynamically provide guidance for the branch selection of each component.

{\bf{Intra-projection:}}
In order to achieve the self-separating of the mixed components and aggregate the patterns of the same frequencies, we extract information within a component. For the input component $\bm{\mathcal{I}}^{i}$, the process is as follows:
\begin{equation}
\begin{split}
    \bm{h}^{i} &= \mathrm{MLP}\left(\bm{\mathcal{I}}^{i}\right) \\
    \bm{k}^{i} &= \mathrm{SoftMax}\left(\bm{h}^{i}\bm{\mathcal{T}}\right), \\
\end{split}
\end{equation}
where $\bm{h}^{i}\in\mathbb{R}^{d}$ represents the hidden states capturing global features of the $i$-th component. $\bm{\mathcal{T}}\in\mathbb{R}^{d\times 2}$ is the transformation matrix of hidden states. $\bm{k}^{i}\in\mathbb{R}^{2}$ is the global guidance weight for the two branches of the $i$-th component. All components share the same projection network, and their global guidance weights form the key of the branch selection, i.e., $\mathbf{K}=[\bm{k}^{1},\bm{k}^{2},\cdots,\bm{k}^{K}]\in\mathbb{R}^{K\times 2}$.

{\bf{Inter-mask:}}
Due to non-stationarity, the statistical properties of time series may exhibit time-varying behavior. Although EMD can be used to handle non-stationary time series, the patterns or frequencies of components obtained by MCD blocks may still change over time, and simply separating the entire components into two branches using global guidance weights is unreliable. Therefore, for all components $\bm{\mathcal{I}}$, we use a guidance mask to account for time-varying as follows:
\begin{equation}
\begin{split}
    \mathbf{M} &= \mathrm{Conv2D}\left(\bm{\mathcal{I}}\right) \\
    \mathbf{Q} &= \bm{\mathcal{I}}\circ\mathbf{M}, \\
\end{split}
\end{equation}
where $\mathbf{M}\in\mathbb{R}^{T\times K}$ is the guidance mask obtained by convolving with $K$ 2D kernels of shape $O\times K$, each corresponding to one component, and $O$ is a hyper-parameter. $\circ$ denotes the Hadamard product. Since the identical time windows of different components are naturally adjacent in the arrangement of the input, the 2D locality can be easily processed by 2D convolution. $\mathbf{Q}\in\mathbb{R}^{T\times K}$ represents the query used for the branch selection of all components, taking into account the above locality and time-varying properties.

\subsubsection{D-R block}
Fig.~\ref{fig:framework}(b) shows the details of the D-R block. The input sequence $\Tilde{\bm{X}}^{l-1}_{i}$ of the $i$-th D-R block at level $l$ is decomposed into multiple components through the MCD block, and then $\mathbf{Q}$ and $\mathbf{K}$ are generated through BGG to reconstruct two sequences. Each reconstructed sequence is computed as a weighted sum of the components $\bm{\mathcal{I}}$, and the weight assigned to $\bm{\mathcal{I}}^{i}$ is computed by the corresponding value of mask $\mathbf{M}$ and key $\mathbf{K}$ adaptively. The reconstruction in the $i$-th D-R block at level $l$ is formulated as follows:
\begin{equation}
\begin{split}
    \mathbf{P} &= \mathbf{Q}\mathbf{K} \\
    \Tilde{\bm{X}}^{l}_{2i-1}, \Tilde{\bm{X}}^{l}_{2i} &= \mathrm{MLP}\left(\mathbf{P}\right),
\end{split}
\end{equation}
where $\mathbf{P}\in\mathbb{R}^{T\times 2}$ represents the reconstructed sequences. $\Tilde{\bm{X}}^{l}_{2i-1}, \Tilde{\bm{X}}^{l}_{2i}$ are the output of the $i$-th D-R block at level $l$.

The aforementioned process separates the different frequency patterns mixed in the same component, and reconstructs them into new sequences based on the weights, which can gather the information of the same frequencies that is previously scattered and mixed in various components. The reconstructed sequences will undergo other rounds of decomposition and reconstruction in subsequent levels. With the stacking of levels, different frequency patterns will be effectively separated, and the patterns of the same frequency will be gathered in the same component.

\subsection{IF Module}
After a D-R-D module with $L$ levels, we obtain $2^{L-1}K$ components $\Tilde{\bm{\mathcal{I}}}\in\mathbb{R}^{T\times 2^{L-1}K}$,  representing different frequency patterns. In order to model potential interactions between these frequency patterns, we introduce interaction learning among multiple components. There are many advanced methods available for modeling interactions between different features or entities, e.g., graph neural networks (GNNs)~\cite{wu2019graph}, self-attention~\cite{vaswani2017attention}, and DeepFM~\cite{chen2022dexdeepfm}. Since we focus on the effective disentanglement of time series patterns, we only use a general graph neural network. We treat each component as a node in a graph, and use basic graph learning techniques to obtain a self-adaptive adjacency matrix. The message passing used to model the interactions among patterns of multiple frequencies is formulated as follows:
\begin{equation}
\begin{split}
    \mathbf{E}_{1} &= \mathrm{Linear}~(\Tilde{\bm{\mathcal{I}}}) \\
    \mathbf{E}_{2} &= \mathrm{Linear}~(\Tilde{\bm{\mathcal{I}}}) \\
    \mathbf{A}_\mathrm{adp} &= \mathrm{SoftMax}\left(\mathrm{ReLU(\mathbf{E}_{1}\mathbf{E}_{2}^\mathrm{T})}\right) \\
    \mathbf{Z}_\mathrm{mid} &= \mathrm{ReLU}\left(\mathbf{W}_\mathrm{G}\mathbf{Z}_\mathrm{in}\mathbf{A}_\mathrm{adp}\right),
\end{split}
\end{equation}
where $\mathbf{E}_{1},\mathbf{E}_{2}\in\mathbb{R}^{d_{\mathrm{em}}\times 2^{L-1}K}$ represent node embeddings, $\mathbf{A}_\mathrm{adp}\in\mathbb{R}^{2^{L-1}K\times 2^{L-1}K}$ is a self-adaptive adjacency matrix, $\mathbf{W}_\mathrm{G}\in\mathbb{R}^{d_\mathrm{mid}\times d_\mathrm{in}}$ is the weight of the graph convolution, $\mathbf{Z}_\mathrm{in}\in\mathbb{R}^{d_\mathrm{in}\times 2^{L-1}K}$ is the input of the GNN, generated through the linear transformation of $\Tilde{\bm{\mathcal{I}}}$, and $\mathbf{Z}_\mathrm{mid}\in\mathbb{R}^{d_\mathrm{mid}\times 2^{L-1}K}$ is the output of the graph convolution.

Similar to multi-channel in convolutional neural networks, multi-graph allows the model to jointly attend to information from different representation subspaces at different graphs, which enriches the capability of the model and stabilizes the training process. Therefore, we incorporate multi-graph into D-PAD, which can be formulated as follows:
\begin{equation}
\begin{split}
    \mathbf{Z}^{\prime}_\mathrm{mid} &= \mathrm{Concat}\left(\mathbf{Z}^{1}_\mathrm{mid},\cdots,\mathbf{Z}^{M}_\mathrm{mid}\right)+\mathbf{Z}_\mathrm{in} \\
    \mathbf{Z}_\mathrm{out} &= \mathrm{Sum}\left(\mathrm{Linear}~(\mathbf{Z}^{\prime}_\mathrm{mid})\right),
\end{split}
\end{equation}
where $\mathbf{Z}^{i}_\mathrm{mid}\in\mathbb{R}^{d_\mathrm{mid}\times 2^{L-1}K}$ is the output of the $j$-th graph in $M$ graphs, $\mathbf{Z}^{\prime}_\mathrm{mid}$ is composed of the output of the GNN and skip-connection, and $\mathbf{Z}_\mathrm{out}\in\mathbb{R}^{d_\mathrm{out}}$ is the fusion of features corresponding to all components.

\subsection{Forecasting}

We use an MLP to predict future values of length $H$ as follows:
\begin{equation}
    \bm{\hat{X}} = \mathrm{MLP}\left(\mathbf{Z}_\mathrm{out}\right).
\end{equation}

The loss function is defined as follows:
\begin{equation}
    \mathcal{L}(\mathbf{\Phi}) = \frac{1}{H}{\sum_{t=1}^{H}\left\| {x_{t} - {\hat{x}}_{t}} \right\|},
\end{equation}
where $\mathbf{\Phi}$ denotes all the learnable parameters in D-PAD. $x_{t}$ and $\hat{x}_{t}$ are the ground truth and the forecasting results, respectively.

\begin{table*}[tbp]
    \caption{Results of multivariate long-term forecasting. The best results are in bold and the second best are underlined. IMP shows the improvement of D-PAD over the best baseline.}\label{tab:Results_M}
    \centering
    \begin{threeparttable}
    \renewcommand{\multirowsetup}{\centering}
    \setlength{\tabcolsep}{3.5pt}
    \begin{tabular}{c|c|cc||cccccccccccccccc}
        \toprule
        \multicolumn{2}{c|}{\multirow{2}{*}{Methods}} & 
        \multicolumn{2}{c||}{\textbf{D-PAD}} & 
        \multicolumn{2}{c|}{DLinear} & 
        \multicolumn{2}{c|}{*N-HITS} & 
        \multicolumn{2}{c|}{*LaST} & 
        \multicolumn{2}{c|}{*SCINet} & 
        \multicolumn{2}{c|}{FEDformer} & 
        \multicolumn{2}{c|}{Autoformer} & 
        \multicolumn{2}{c|}{Informer} & 
        \multicolumn{2}{c}{\multirow{2}{*}{IMP}} \\

        \multicolumn{2}{c|}{} & 
        \multicolumn{2}{c||}{\textbf{(Ours)}} & 
        \multicolumn{2}{c|}{(AAAI 2023)} & 
        \multicolumn{2}{c|}{(AAAI 2023)} & 
        \multicolumn{2}{c|}{(NIPS 2022)} & 
        \multicolumn{2}{c|}{(NIPS 2022)} & 
        \multicolumn{2}{c|}{(ICML 2022)} & 
        \multicolumn{2}{c|}{(NIPS 2021)} & 
        \multicolumn{2}{c|}{(AAAI 2021)} & 
        \multicolumn{2}{c}{} \\

        \midrule
        \multicolumn{2}{c|}{Metrics} & 
        \multicolumn{1}{c}{MSE} & \multicolumn{1}{c||}{MAE} & 
        \multicolumn{1}{c}{MSE} & \multicolumn{1}{c}{MAE} & 
        \multicolumn{1}{c}{MSE} & \multicolumn{1}{c}{MAE} & 
        \multicolumn{1}{c}{MSE} & \multicolumn{1}{c}{MAE} & 
        \multicolumn{1}{c}{MSE} & \multicolumn{1}{c}{MAE} & 
        \multicolumn{1}{c}{MSE} & \multicolumn{1}{c}{MAE} & 
        \multicolumn{1}{c}{MSE} & \multicolumn{1}{c}{MAE} & 
        \multicolumn{1}{c}{MSE} & \multicolumn{1}{c}{MAE} \\
        
    \midrule
    \multirow{4}{*}{\rotatebox{90}{ETTh1}} & 
    96    & \textbf{0.357} & \textbf{0.376} & \underline{0.375} & \underline{0.399} & 0.475 & 0.498 & 0.395 & 0.407 & 0.401 & 0.400 & 0.376 & 0.415 & 0.435 & 0.446 & 0.941 & 0.769 & 4.80\% & 5.76\% \\
    & 192   & \textbf{0.394} & \textbf{0.402} & \underline{0.405} & \underline{0.416} & 0.492 & 0.519 & 0.463 & 0.456 & 0.468 & 0.457 & 0.423 & 0.446 & 0.456 & 0.457 & 1.007 & 0.786 & 2.72\% & 3.37\% \\
    & 336   & \textbf{0.374} & \textbf{0.406} & \underline{0.439} & \underline{0.443} & 0.550 & 0.564 & 0.556 & 0.502  & 0.516 & 0.509 & 0.444 & 0.462 & 0.486 & 0.487 & 1.038 & 0.784 & 14.81\% & 8.35\% \\
    & 720   & \textbf{0.419} & \textbf{0.442} & 0.472 & \underline{0.490} & 0.598 & 0.641 & 0.714 & 0.612 & 0.554 & 0.535 & \underline{0.469} & 0.492 & 0.515 & 0.517 & 1.144 & 0.857 & 10.66\% & 9.80\% \\
    \midrule
    \multirow{4}{*}{\rotatebox{90}{ETTh2}} & 
    96    & \textbf{0.270} & \textbf{0.327} & \underline{0.289} & \underline{0.353} & 0.328 & 0.364 & 0.313 & 0.372 & 0.315 & 0.359 & 0.332 & 0.374 & 0.332 & 0.368 & 1.549 & 0.952 & 6.57\% & 7.37\% \\
    & 192   & \textbf{0.331} & \textbf{0.368} & 0.383 & 0.418 & \underline{0.372} & \underline{0.408} & 0.519 & 0.554 & 0.402 & 0.425 & 0.407 & 0.443 & 0.426 & 0.434 & 3.792 & 1.542 & 11.02\% & 9.80\% \\
    & 336   & \textbf{0.321} & \textbf{0.370} & 0.448 & 0.465 & \underline{0.397} & \underline{0.421} & 0.722 & 0.684 & 0.414 & 0.437 & 0.400 & 0.447 & 0.477 & 0.479 & 4.215 & 1.642 & 19.14\% & 12.11\% \\
    & 720   & \textbf{0.369} & \textbf{0.415} & 0.605 & 0.551 & 0.461 & 0.497 & 0.817 & 0.741 & 0.492 & 0.513 & \underline{0.412} & \underline{0.469} & 0.453 & 0.490 & 3.656 & 1.619 & 10.44\% & 11.51\% \\
    \midrule
    \multirow{4}{*}{\rotatebox{90}{ETTm1}} & 
    96    & \textbf{0.285} & \textbf{0.328} & \underline{0.299} & \underline{0.343} & 0.370 & 0.468 & 0.306 & 0.349 & 0.305 & 0.352 & 0.326 & 0.390 & 0.510 & 0.492 & 0.626 & 0.560 & 4.68\% & 4.37\% \\
    & 192   & \textbf{0.323} & \textbf{0.349} & \underline{0.335} & \underline{0.365} & 0.436 & 0.488 & 0.349 & 0.373  & 0.353 & 0.371 & 0.365 & 0.415 & 0.514 & 0.495 & 0.725 & 0.619 & 3.58\% & 4.38\% \\
    & 336   & \textbf{0.351} & \textbf{0.372} & \underline{0.369} & \underline{0.386} & 0.483 & 0.510 & 0.389 & 0.400  & 0.387 & 0.404 & 0.392 & 0.425 & 0.510 & 0.492 & 1.005 & 0.741 & 4.88\% & 3.63\% \\
    & 720   & \textbf{0.412} & \textbf{0.405} & \underline{0.425} & \underline{0.421} & 0.489 & 0.537 & 0.480 & 0.459  & 0.449 & 0.451 & 0.446 & 0.458 & 0.527 & 0.493 & 1.133 & 0.845 & 3.06\% & 3.80\% \\
    \midrule
    \multirow{4}{*}{\rotatebox{90}{ETTm2}} & 
    96    & \textbf{0.162} & \textbf{0.247} & \underline{0.167} & \underline{0.260} & 0.184 & 0.262 & 0.170 & 0.262 & 0.179 & 0.280 & 0.180 & 0.271 & 0.205 & 0.293 & 0.355 & 0.462 & 2.99\% & 5.00\% \\
    & 192   & \textbf{0.218} & \textbf{0.283} & \underline{0.224} & 0.303 & 0.260 & \underline{0.293} & 0.229 & 0.308  & 0.244 & 0.304 & 0.252 & 0.318 & 0.278 & 0.336 & 0.595 & 0.586 & 2.68\% & 3.41\% \\
    & 336   & \textbf{0.267} & \textbf{0.321} & \underline{0.281} & \underline{0.342} & 0.313 & 0.359 & 0.326 & 0.382  & 0.318 & 0.355 & 0.324 & 0.364 & 0.343 & 0.379 & 1.270 & 0.871 & 4.98\% & 6.14\% \\
    & 720   & \textbf{0.353} & \textbf{0.372} & \underline{0.397} & 0.421 & 0.411 & 0.421 & 0.863 & 0.651 & 0.413 & 0.432 & 0.410 & 0.420 & 0.414 & \underline{0.419} & 3.001 & 1.267 & 11.08\% & 11.22\% \\
    \midrule
    \multirow{4}{*}{\rotatebox{90}{Electricity}} & 
    96    & \textbf{0.128} & \textbf{0.218} & \underline{0.140} & \underline{0.237} & 0.151 & 0.254 & 0.145 & 0.238 & 0.169 & 0.257 & 0.186 & 0.302 & 0.196 & 0.313 & 0.304 & 0.393 & 8.57\% & 8.02\% \\
    & 192 & \textbf{0.142} & \textbf{0.233} & \underline{0.153} & \underline{0.249} & 0.170 & 0.273 & 0.159 & \underline{0.249} & 0.183 & 0.270 & 0.197 & 0.311 & 0.211 & 0.324 & 0.327 & 0.417 & 7.19\% & 6.43\% \\
    & 336 & \textbf{0.161} & \textbf{0.254} & \underline{0.169} & \underline{0.267} & 0.200 & 0.291 & 0.183 & 0.278 & 0.192 & 0.283 & 0.213 & 0.328 & 0.214 & 0.327 & 0.333 & 0.422 & 4.73\% & 4.87\% \\
    & 720 & \textbf{0.190} & \textbf{0.282} & \underline{0.203} & \underline{0.301} & 0.244 & 0.356 & 0.221 & 0.304 & 0.234 & 0.322 & 0.233 & 0.344 & 0.236 & 0.342 & 0.351 & 0.427 & 6.40\% & 6.31\% \\
    \midrule
    \multirow{4}{*}{\rotatebox{90}{Traffic}} & 
    96    & \textbf{0.359} & \textbf{0.236} & 0.410 & \underline{0.282} & \underline{0.407} & 0.290 & 0.694 & 0.375 & 0.625 & 0.407 & 0.576 & 0.359 & 0.597 & 0.371 & 0.733 & 0.410 & 11.79\% & 16.31\% \\
    & 192   & \textbf{0.377} & \textbf{0.245} & \underline{0.423} & \underline{0.287} & \underline{0.423} & 0.302 & 0.647 & 0.354 & 0.549 & 0.364 & 0.610 & 0.380 & 0.607 & 0.382 & 0.777 & 0.435 & 10.87\% & 14.63\% \\
    & 336   & \textbf{0.391} & \textbf{0.253} & \underline{0.436} & \underline{0.296} & 0.446 & 0.321 & 0.650 & 0.355 & 0.557 & 0.371 & 0.608 & 0.375 & 0.623 & 0.387 & 0.776 & 0.434 & 10.32\% & 14.53\% \\
    & 720   & \textbf{0.413} & \textbf{0.272} & \underline{0.466} & \underline{0.315} & 0.528 & 0.369 & 0.683 & 0.375 & 0.626 & 0.398 & 0.621 & 0.375 & 0.639 & 0.395 & 0.827 & 0.466 & 11.37\% & 13.65\% \\
    \midrule
    \multirow{4}{*}{\rotatebox{90}{Weather}} & 
    96    & \textbf{0.143} & \textbf{0.181} & 0.176 & 0.237 & \underline{0.160} & \underline{0.197} & 0.166 & 0.218 & 0.243 & 0.318 & 0.238 & 0.314 & 0.249 & 0.329 & 0.354 & 0.405 & 10.63\% & 8.12\% \\
    & 192 & \textbf{0.189} & \textbf{0.229} & 0.220 & 0.282 & 0.207 & 0.265 & \underline{0.204} & \underline{0.247} & 0.281 & 0.329 & 0.275 & 0.329 & 0.325 & 0.370 & 0.419 & 0.434 & 7.35\% & 7.29\% \\
    & 336 & \textbf{0.239} & \textbf{0.268} & 0.265 & 0.319 & 0.273 & 0.301 & \underline{0.252} & \underline{0.284} & 0.337 & 0.371 & 0.339 & 0.377 & 0.351 & 0.391 & 0.583 & 0.543 & 5.16\% & 5.63\% \\
    & 720 & \textbf{0.304} & \textbf{0.313} & 0.323 & 0.362 & 0.363 & 0.352 & \underline{0.315} & \underline{0.325} & 0.392 & 0.413 & 0.389 & 0.409 & 0.415 & 0.426 & 0.916 & 0.705 & 3.49\% & 3.69\% \\
    \bottomrule
    \end{tabular}
    \begin{tablenotes}
        \scriptsize
        \item * denotes method run with an input length of 336 and default parameters.
    \end{tablenotes}
    \end{threeparttable}
\end{table*}

\section{Experiments}
\subsection{Datasets and Settings}
{\bf{Datasets.}} We evaluate D-PAD for multivariate forecasting on seven real-world datasets: Electricity Transformer Temperature (\textit{ETTh1}, \textit{ETTh2}, \textit{ETTm1}, and \textit{ETTm2})~\cite{zhou2021informer}, \textit{Electricity}\footnote{https://github.com/laiguokun/multivariate-time-series-data/\label{multivariate-ts-data}}, \textit{Traffic}\textsuperscript{\ref {multivariate-ts-data}}, and \textit{Weather}\footnote{https://www.bgc-jena.mpg.de/wetter/}, and for univariate forecasting on the first four datasets following the previous works~\cite{zeng2023transformers, nie2022time, wang2022learning}. The details of the seven datasets are given as follows:
\begin{itemize}
    \item \textit{ETT}~\cite{zhou2021informer} captures the electricity transformer temperature, recorded hourly, i.e., \textit{ETTh1} and \textit{ETTh2}, and every 15 minutes, i.e., \textit{ETTm1} and \textit{ETTm2}, over two years, each of them contains 7 oil and load features of electricity transformers from July 2016 to July 2018.
    \item \textit{Electricity} records the hourly electricity consumption of 321 clients from 2012 to 2014.
    \item \textit{Traffic} collects hourly data that describe the road occupancy rates measured by 862 sensors on San Francisco Bay area freeways.
    \item \textit{Weather} includes meteorological time series with 21 weather indicators collected every 10 minutes from the Weather Station of the Max Planck Biogeochemistry Institute in 2020.
\end{itemize}

We follow the standard protocol and split all datasets into training, validation, and test set in chronological order by the ratio of 6:2:2 for \textit{ETT} datasets and 7:1:2 for the others.

{\bf{Settings.}}
The source code of D-PAD is available at GitHub\footnote{https://github.com/XYBbo5/D-PAD/}. D-PAD is implemented in Python with PyTorch 1.9.0 and trained on 4 NVIDIA GeForce RTX 3080 Ti GPU cards using Adam optimizer with an initial learning rate of 0.0001 and a batch size of 32. The training process is early stopped when there is no improvement within 5 epochs. RevIN~\cite{kim2021reversible} is used to help mitigating the distribution shift effect.

By default, D-PAD employs a 2-level D-R-D architecture with a graph number of $M=1$ and a hidden dimension of $d_\mathrm{em}=d_\mathrm{in}=d_\mathrm{mid}=d_\mathrm{out}=256$ for \textit{electricity} and \textit{traffic} datasets, and 336 for other datasets. An MCD block decomposes the time series into $K=6$ components. The length of the SE kernel and 2D convolution kernel is set to the shortest length of 3, i.e., $2C+1=3$ and $O=3$.

Mean Square Error (MSE) and Mean Absolute Error (MAE) are exploited as evaluation metrics, which are defined as follows:
\begin{equation}
\begin{split}
    \mathrm{MSE} &= \frac{1}{H}{\sum_{t=1}^{H}\left\| {x_{t} - {\hat{x}}_{t}} \right\|^{2}}, \\
    \mathrm{MAE} &= \frac{1}{H}{\sum_{t=1}^{H}\left\| {x_{t} - {\hat{x}}_{t}} \right\|}. \\
\end{split}
\end{equation}

\subsection{Methods for Comparison}
We compared D-PAD with the state-of-the-art (SOTA) baselines that emphasize temporal patterns disentangling. The details of methods are as follows:

{\bf{Seasonal-trend decomposition methods:}}
\begin{itemize}
    \item DLinear~\cite{zeng2023transformers}: It stands for seasonal-trend decomposition models that integrate with linear layers, maintaining a simple structure.
    \item LaST~\cite{wang2022learning}: It stands for models employing variational inference in designing trend and seasonal representation learning and disentanglement mechanisms.
    \item Autoformer~\cite{wu2021autoformer}: It stands for the Transformer-based models with seasonal-trend decomposition that make series decomposition as basic inner blocks.
\end{itemize}

{\bf{Multi-component decomposition methods:}}
\begin{itemize}
    \item N-HITS~\cite{challu2023nhits}: It stands for multi-component decomposition models that employ deep stacked architectures composed of fully connected layers.
    \item FEDformer~\cite{zhou2022fedformer}: It stands for the Transformer-based models with multi-component decomposition that incorporate frequency domain analysis techniques.
\end{itemize}

{\bf{Methods without decomposition:}}
\begin{itemize}
    \item Informer~\cite{zhou2021informer}: It stands for the SOTA time series forecasting models without decomposition.
    \item SCINet~\cite{liu2022scinet}: It stands for the multi-level time series forecasting models that model time series with complex temporal dynamics.
\end{itemize}

To ensure a fair comparison, we adopt the same settings as the original publications~\cite{zeng2023transformers, nie2022time}. Specifically, we evaluate D-PAD on each dataset with an input length $T=336$ and prediction length $H\in\{96,192,336,720\}$. We refer to the baseline results from~\cite{nie2022time}. For baselines not covered in~\cite{nie2022time}, we run them using an input length of 336 and default parameters with their publicly available codes.

\begin{table*}[tbp]
    \vspace{-10pt}
  \caption{Results of univariate long-term forecasting on \textit{ETT} datasets. The best results are in bold and the second best are underlined. IMP shows the improvement of D-PAD over the best baseline.}\label{tab:Results_S}
    \centering
    \begin{threeparttable}
    \renewcommand{\multirowsetup}{\centering}
    \setlength{\tabcolsep}{3.5pt}
    \begin{tabular}{c|c|cc||cccccccccccccccc}
        \toprule
        \multicolumn{2}{c|}{\multirow{2}{*}{Methods}} & 
        \multicolumn{2}{c||}{\textbf{D-PAD}} & 
        \multicolumn{2}{c|}{DLinear} & 
        \multicolumn{2}{c|}{*N-HITS} & 
        \multicolumn{2}{c|}{*LaST} & 
        \multicolumn{2}{c|}{*SCINet} & 
        \multicolumn{2}{c|}{FEDformer} & 
        \multicolumn{2}{c|}{Autoformer} & 
        \multicolumn{2}{c|}{Informer} & 
        \multicolumn{2}{c}{\multirow{2}{*}{IMP}} \\

        \multicolumn{2}{c|}{} & 
        \multicolumn{2}{c||}{\textbf{(Ours)}} & 
        \multicolumn{2}{c|}{(AAAI 2023)} & 
        \multicolumn{2}{c|}{(AAAI 2023)} & 
        \multicolumn{2}{c|}{(NIPS 2022)} & 
        \multicolumn{2}{c|}{(NIPS 2022)} & 
        \multicolumn{2}{c|}{(ICML 2022)} & 
        \multicolumn{2}{c|}{(NIPS 2021)} & 
        \multicolumn{2}{c|}{(AAAI 2021)} & 
        \multicolumn{2}{c}{} \\

        \midrule
        \multicolumn{2}{c|}{Metrics} & 
        \multicolumn{1}{c}{MSE} & \multicolumn{1}{c||}{MAE} & 
        \multicolumn{1}{c}{MSE} & \multicolumn{1}{c}{MAE} & 
        \multicolumn{1}{c}{MSE} & \multicolumn{1}{c}{MAE} & 
        \multicolumn{1}{c}{MSE} & \multicolumn{1}{c}{MAE} & 
        \multicolumn{1}{c}{MSE} & \multicolumn{1}{c}{MAE} & 
        \multicolumn{1}{c}{MSE} & \multicolumn{1}{c}{MAE} & 
        \multicolumn{1}{c}{MSE} & \multicolumn{1}{c}{MAE} & 
        \multicolumn{1}{c}{MSE} & \multicolumn{1}{c}{MAE} & 
        \multicolumn{1}{c}{MSE} & \multicolumn{1}{c}{MAE} \\
        
    \midrule
    \multirow{4}{*}{\rotatebox{90}{ETTh1}} & 
    96    & \textbf{0.052} & \textbf{0.171} & \underline{0.056} & \underline{0.180} & 0.068 & 0.184 & 0.058 & 0.184 & 0.062 & 0.188 & 0.079 & 0.215 & 0.071 & 0.206 & 0.193 & 0.377 & 7.14\% & 5.00\% \\
    & 192   & \textbf{0.068} & \textbf{0.194} & \underline{0.071} & \underline{0.204} & 0.086 & 0.231 & 0.079 & 0.215 & 0.089 & 0.225 & 0.104 & 0.245 & 0.114 & 0.262 & 0.217 & 0.395 & 4.23\% & 4.90\% \\
    & 336   & \textbf{0.077} & \textbf{0.219} & 0.098 & 0.244 & 0.097 & 0.249 & 0.097 & 0.243 & \underline{0.091} & \underline{0.238} & 0.119 & 0.270 & 0.107 & 0.257 & 0.202 & 0.381 & 15.38\% & 7.98\% \\
    & 720   & \textbf{0.085} & \textbf{0.230} & 0.189 & 0.359 & 0.152 & 0.318 & 0.193 & 0.366 & 0.166 & 0.333 & 0.142 & 0.299 & \underline{0.126} & \underline{0.283} & 0.183 & 0.355 & 32.54\% & 18.73\% \\
    \midrule
    \multirow{4}{*}{\rotatebox{90}{ETTh2}} & 
    96    & \textbf{0.115} & \textbf{0.262} & 0.131 & 0.279 & 0.134 & 0.289 & 0.135 & 0.283 & 0.133 & 0.283 & \underline{0.128} & \underline{0.271} & 0.153 & 0.306 & 0.213 & 0.313 & 10.16\% & 3.32\% \\
    & 192   & \textbf{0.148} & \textbf{0.308} & 0.176 & 0.329 & \underline{0.166} & 0.325 & 0.176 & 0.330 & \underline{0.166} & \underline{0.321} & 0.185 & 0.330 & 0.246 & 0.351 & 0.227 & 0.387 & 10.84\% & 4.05\% \\
    & 336   & \textbf{0.152} & \textbf{0.319} & 0.209 & 0.367 & 0.204 & 0.356 & 0.204 & 0.363 & \underline{0.179} & \underline{0.340} & 0.231 & 0.378 & 0.246 & 0.389 & 0.424 & 0.401 & 15.08\% & 6.18\% \\
    & 720   & \textbf{0.198} & \textbf{0.360} & 0.276 & 0.426 & 0.264 & \underline{0.405} & \underline{0.255} & 0.408 & 0.256 & 0.409 & 0.278 & 0.420 & 0.268 & 0.409 & 0.291 & 0.439 & 22.35\% & 11.11\% \\
    \midrule
    \multirow{4}{*}{\rotatebox{90}{ETTm1}} & 
    96    & \textbf{0.024} & \textbf{0.118} & \underline{0.028} & \underline{0.123} & 0.032 & 0.127 & 0.037 & 0.144 & \underline{0.028} & 0.125 & 0.033 & 0.140 & 0.056 & 0.183 & 0.109 & 0.277 & 14.29\% & 4.07\% \\
    & 192   & \textbf{0.037} & \textbf{0.148} & 0.045 & \underline{0.156} & \underline{0.043} & 0.164 & 0.056 & 0.176 & 0.047 & 0.163 & 0.058 & 0.186 & 0.081 & 0.216 & 0.151 & 0.310 & 13.95\% & 5.13\% \\
    & 336   & \textbf{0.053} & \textbf{0.170} & \underline{0.061} & \underline{0.182} & 0.088 & 0.184 & 0.083 & 0.216 & 0.105 & 0.250 & 0.084 & 0.231 & 0.076 & 0.218 & 0.427 & 0.591 & 13.11\% & 6.59\% \\
    & 720   & \textbf{0.072} & \textbf{0.203} & \underline{0.080} & \underline{0.210} & 0.101 & 0.234 & 0.092 & 0.227 & 0.088 & 0.224 & 0.102 & 0.250 & 0.110 & 0.267 & 0.438 & 0.586 & 10.00\% & 3.33\% \\
    \midrule
    \multirow{4}{*}{\rotatebox{90}{ETTm2}} & 
    96    & \textbf{0.059} & \textbf{0.176} & \underline{0.063} & \underline{0.183} & 0.068 & 0.188 & 0.067 & 0.189 & 0.066 & 0.187 & 0.067 & 0.198 & 0.065 & 0.189 & 0.088 & 0.225 & 6.35\% & 3.83\% \\
    & 192   & \textbf{0.082} & \textbf{0.214} & 0.092 & 0.227 & 0.091 & 0.231 & 0.095 & 0.231 & \underline{0.088} & \underline{0.222} & 0.102 & 0.245 & 0.118 & 0.256 & 0.132 & 0.283 & 6.82\% & 3.60\% \\
    & 336   & \textbf{0.104} & \textbf{0.249} & \underline{0.119} & \underline{0.261} & 0.125 & 0.283 & 0.120 & 0.262 & 0.145 & 0.294 & 0.130 & 0.279 & 0.154 & 0.305 & 0.180 & 0.336 & 12.61\% & 4.60\% \\
    & 720   & \textbf{0.155} & \textbf{0.307} & 0.175 & 0.320 & 0.174 & 0.329 & 0.174 & 0.321 & \underline{0.165} & \underline{0.315} & 0.178 & 0.325 & 0.182 & 0.335 & 0.300 & 0.435 & 6.06\% & 2.54\% \\
    \bottomrule
    \end{tabular}
    \begin{tablenotes}
        \scriptsize
        \item * denotes method run with an input length of 336 and default parameters.
    \end{tablenotes}
    \end{threeparttable}
    \vspace{-8pt}
\end{table*}

\begin{table}[tbp]
\vspace{-5pt}
\renewcommand{\arraystretch}{0.9}
  \centering
  \caption{Results of D-PAD and its variants about MCD block and BGG on \textit{ETT} datasets. The best results are in bold.} \label{tab:ablation_1}
    \setlength{\tabcolsep}{4.3pt}
    \begin{tabular}{c|c|cc|cc|cc|cc}
    \toprule
    \multicolumn{2}{c|}{\multirow{2}[2]{*}{Variants}} & 
    \multicolumn{2}{c|}{ETTh1} & 
    \multicolumn{2}{c|}{ETTh2} & 
    \multicolumn{2}{c|}{ETTm1} & 
    \multicolumn{2}{c}{ETTm2} \\
    \cmidrule{3-10}    
    \multicolumn{2}{c|}{} & 
    \multicolumn{1}{c}{MSE} & 
    \multicolumn{1}{c|}{MAE} & 
    \multicolumn{1}{c}{MSE} & 
    \multicolumn{1}{c|}{MAE} & 
    \multicolumn{1}{c}{MSE} & 
    \multicolumn{1}{c|}{MAE} & 
    \multicolumn{1}{c}{MSE} & 
    \multicolumn{1}{c}{MAE} \\
    \midrule
    \multicolumn{1}{c|}{\multirow{4}{*}{\rotatebox{90}{D-PAD}}} & 96    & \textbf{0.357} & \textbf{0.376} & \textbf{0.270} & \textbf{0.327} & \textbf{0.285} & \textbf{0.328} & \textbf{0.162} & \textbf{0.247} \\
          & 192   & \textbf{0.394} & \textbf{0.402} & \textbf{0.331} & \textbf{0.368} & \textbf{0.323} & \textbf{0.349} & \textbf{0.218} & \textbf{0.283} \\
          & 336   & \textbf{0.374} & \textbf{0.406} & \textbf{0.321} & \textbf{0.370} & \textbf{0.351} & \textbf{0.372} & \textbf{0.267} & \textbf{0.321} \\
          & 720   & \textbf{0.419} & \textbf{0.442} & \textbf{0.369} & \textbf{0.415} & \textbf{0.412} & \textbf{0.405} & \textbf{0.353} & \textbf{0.372} \\
    \midrule
    \multicolumn{1}{c|}{\multirow{4}{*}{\rotatebox{90}{D-PAD-L}}} & 96    & 0.397 & 0.408 & 0.332 & 0.372 & 0.296 & 0.337 & 0.166 & 0.251 \\
          & 192   & 0.410 & 0.416 & 0.348 & 0.391 & 0.338 & 0.358 & 0.279 & 0.361 \\
          & 336   & 0.421 & 0.431 & 0.367 & 0.410 & 0.375 & 0.384 & 0.284 & 0.339 \\
          & 720   & 0.469 & 0.488 & 0.539 & 0.515 & 0.433 & 0.420 & 0.378 & 0.400 \\
    \midrule
    \multicolumn{1}{c|}{\multirow{4}{*}{\rotatebox{90}{D-PAD-F}}} & 96    & 0.392 & 0.401 & 0.314 & 0.356 & 0.301 & 0.338 & 0.179 & 0.264 \\
          & 192   & 0.411 & 0.415 & 0.355 & 0.384 & 0.345 & 0.360 & 0.274 & 0.354 \\
          & 336   & 0.418 & 0.432 & 0.374 & 0.422 & 0.374 & 0.389 & 0.286 & 0.338 \\
          & 720   & 0.479 & 0.498 & 0.527 & 0.530 & 0.448 & 0.439 & 0.384 & 0.411 \\
    \midrule
    \multicolumn{1}{c|}{\multirow{4}{*}{\rotatebox{90}{D-PAD-H}}} & 96    & 0.371 & 0.393 & 0.279 & 0.343 & 0.310 & 0.339 & 0.172 & 0.255 \\
          & 192   & 0.433 & 0.423 & 0.338 & 0.379 & 0.335 & 0.351 & 0.227 & 0.290 \\
          & 336   & 0.384 & 0.411 & 0.326 & 0.380 & 0.357 & 0.385 & 0.276 & 0.331 \\
          & 720   & 0.426 & 0.459 & 0.403 & 0.436 & 0.422 & 0.416 & 0.355 & 0.374 \\
    \bottomrule
    \end{tabular}%
    \vspace{-8pt}
\end{table}%

\subsection{Main Results}

Table~\ref{tab:Results_M} and Table~\ref{tab:Results_S} summarize the results of multivariate and univariate forecasting, respectively. The following phenomena can be observed:
\begin{itemize}
    \item D-PAD achieves the consistent SOTA performance and outperforms all baselines on all datasets and all prediction length settings. Quantitatively, in the multivariate setting, D-PAD surpasses the best baseline by an average of 9.48\% and 7.15\% in MSE and MAE, respectively. In the univariate setting, the improvement is 12.56\% and 5.93\% for MSE and MAE, respectively. It demonstrates the effectiveness of D-PAD in modeling intricate temporal patterns and indicates its wide applicability and stability across various domains and prediction horizons.
    
    \item The baselines that incorporate the decomposition into deep models, i.e., DLinear and LaST, significantly outperform the pure deep model, i.e., Informer. This is because decomposition methods simplify complex data into more manageable components, which is more conducive to subsequent analysis.
    
    \item Multiple components decomposition models, i.e., FEDformer and N-HITS, outperform seasonal-trend decomposition model, i.e., Autoformer. This is because they effectively disentangle multiple intricate patterns into various components conducive to analysis. However, these models underperform in certain cases due to limitations in time domain representation.
\end{itemize}

\subsection{Ablation Study}
\label{subsec:ablation}

To evaluate the impact of each main component used in D-PAD, we conduct the ablation study on \textit{ETT} datasets.

{\bf{MCD block: }}
To investigate the effect of the MCD block, we compare D-PAD with two variants. The detailed descriptions of variants are as follows:
\begin{itemize}
    \item D-PAD-L: It replaces the MCD block with a learning-based decomposition approach. An MLP is used to generate multiple components under a constraint of the loss between the reconstructed and original sequences, as the reconstructed sequence should preserve the original patterns as closely as possible.

    \item D-PAD-F: It replaces the MEMD in MCD block with the Discrete Fourier Transform and selects $K$ components with the highest weights. D-PAD-F achieves multi-component decomposition with frequency domain analysis methods.
\end{itemize}

The results are shown in the Table~\ref{tab:ablation_1}, from which we can observe the following phenomena:

1) D-PAD outperforms D-PAD-L, indicating the superiority of the MCD block over learning-based decomposition methods. The inherent inductive bias of MCD block ensures effective disentanglement by decomposing series into multiple components with distinct frequency ranges. While learning-based decomposition methods, constrained only by the reconstruction loss, suffer from the issue of information mixing.

2) D-PAD outperforms D-PAD-F, indicating that MEMD is superior to Fourier decomposition. This is because, compared to frequency domain analysis methods, MEMD can effectively handle the nonlinearity and non-stationarity in time series while exhibiting strong adaptability.

{\bf{BGG: }}
To demonstrate the effectiveness of BGG, we compare D-PAD with the following variant:

\begin{itemize}
    \item D-PAD-H: It replaces BGG with hard selection, discretizing the branch selection of each component. It can be formulated as follows: 

\begin{equation}
\begin{split}
    \bm{z}^{j} &= \mathrm{MLP}\left(\bm{\mathcal{I}}^{j}\right) \\
    \bm{u}^{j} &= \mathrm{Gumbel\mbox{-}SoftMax} \left(\bm{z}^{j}\right) \\
    \bm{\Tilde{X}}^{l}_{2i-1},\bm{\Tilde{X}}^{l}_{2i} &= \mathrm{MLP}~({\sum_{j = 1}^{K}{\bm{\mathcal{I}}^{j}\bm{u}^{j}}}),
\end{split}
\end{equation}

\end{itemize}
where $\bm{u}^{j}=[1,0]$ or $[0,1]$ is generated by the Gumbel-SoftMax~\cite{jang2016categorical} with hidden state $\bm{z}^{j}\in\mathbb{R}^{2}$. $\bm{\Tilde{X}}^{l}_{2i-1}$ and $\bm{\Tilde{X}}^{l}_{2i}$ are the output of the $i$-th D-R block at level $l$. The results are shown in Table~\ref{tab:ablation_1}, from which we can observe that D-PAD outperforms D-PAD-H. It indicates the effectiveness of the intra-projection and inter-mask in BGG. They enhance disentanglement by guiding the selection of components and considering dependencies both within components and among different components.

{\bf{IF module: }}
To illustrate the effect of component interaction within the IF module, we compare D-PAD with a variant as follows:
\begin{itemize}
    \item D-PAD-W: It removes the IF module. The output of the D-R-D module is directly used for fusion and prediction.
\end{itemize}

The results of the variant experiments and the best baseline are shown in Table~\ref{tab:ablation_IF}, where the best baseline results are composed of the optimal results of each comparative baseline on the \textit{ETT} datasets. We can observe the following phenomena:

1) D-PAD achieves the best performance across all cases, which shows the advantage of introducing interaction learning among multiple components. This is because the IF module effectively models the correlations among multiple mixed temporal patterns in time series.
    
2) Even without the IF module, D-PAD-W still outperforms baselines in most cases, which indicates that the D-R-D module is critical, as intricate temporal patterns are effectively disentangled by the D-R-D module.

\begin{table}[tbp]
\renewcommand{\arraystretch}{0.95}
  \centering
  \caption{Results of D-PAD, D-PAD-W and best baseline on \textit{ETT} datasets. The best results are in bold and the second best are underlined.}\label{tab:ablation_IF}
  \centering
    \renewcommand{\multirowsetup}{\centering}
    \setlength{\tabcolsep}{4pt}
    \begin{tabular}{c|c|cc|cc|cc|cc}
    \toprule
    \multicolumn{2}{c|}{\multirow{2}[2]{*}{Variants}} & 
    \multicolumn{2}{c|}{ETTh1} & 
    \multicolumn{2}{c|}{ETTh2} & 
    \multicolumn{2}{c|}{ETTm1} & 
    \multicolumn{2}{c}{ETTm2} \\
    \cmidrule{3-10}    
    \multicolumn{2}{c|}{} & 
    \multicolumn{1}{c}{MSE} & 
    \multicolumn{1}{c|}{MAE} & 
    \multicolumn{1}{c}{MSE} & 
    \multicolumn{1}{c|}{MAE} & 
    \multicolumn{1}{c}{MSE} & 
    \multicolumn{1}{c|}{MAE} & 
    \multicolumn{1}{c}{MSE} & 
    \multicolumn{1}{c}{MAE} \\
    
    \midrule
    
    \multirow{4}{*}{\rotatebox{90}{D-PAD}} & 96    & \textbf{0.357} & \textbf{0.376} & \textbf{0.270} & \textbf{0.327} & \textbf{0.285} & \textbf{0.328} & \textbf{0.162} & \textbf{0.247} \\
          & 192   & \textbf{0.394} & \textbf{0.402} & \textbf{0.331} & \textbf{0.368} & \textbf{0.323} & \textbf{0.349} & \textbf{0.218} & \textbf{0.283} \\
          & 336   & \textbf{0.374} & \textbf{0.406} & \textbf{0.321} & \textbf{0.370} & \textbf{0.351} & \textbf{0.372} & \textbf{0.267} & \textbf{0.321} \\
          & 720   & \textbf{0.419} &\textbf{ 0.442} & \textbf{0.369} & \textbf{0.415} & \textbf{0.412} & \textbf{0.405} & \textbf{0.353} & \textbf{0.372} \\
    \midrule
    \multirow{4}{*}{\rotatebox{90}{D-PAD-W}} & 96    & \underline{0.372} & \underline{0.397} & \underline{0.283} & \underline{0.340} & \underline{0.294} & \underline{0.340} & 0.170 & \underline{0.251} \\
          & 192   & 0.414 & 0.422 & \underline{0.338} & \underline{0.377} & \underline{0.333} & \underline{0.365} & \underline{0.224} & \underline{0.289} \\
          & 336   & \underline{0.386} & \underline{0.417} & \underline{0.325} & \underline{0.373} & \underline{0.354} & \underline{0.385} & \underline{0.280} & \underline{0.323} \\
          & 720   & \underline{0.429} & \underline{0.451} & \underline{0.403} & \underline{0.428} & \underline{0.423} & \underline{0.421} & \underline{0.369} & \underline{0.382} \\
    \midrule
    \multirow{4}{*}{\rotatebox{90}{\makecell{Best \\ baselines}}} & 96    & 0.375 & 0.399 & 0.289 & 0.353 & 0.299 & 0.343 & \underline{0.167} & 0.260 \\
          & 192   & \underline{0.405} & \underline{0.416} & 0.372 & 0.408 & 0.335 & \underline{0.365} & \underline{0.224} & 0.303 \\
          & 336   & 0.439 & 0.443 & 0.397 & 0.421 & 0.369 & 0.386 & 0.281 & 0.342 \\
          & 720   & 0.472 & 0.490 & 0.412 & 0.469 & 0.425 & \underline{0.421} & 0.397 & 0.419 \\
    \bottomrule
    \end{tabular}%
\end{table}%

\subsection{Parameter Sensitivity Analysis}

\begin{figure}[tbp]
\begin{center}
	\centerline{\includegraphics[width=\columnwidth]{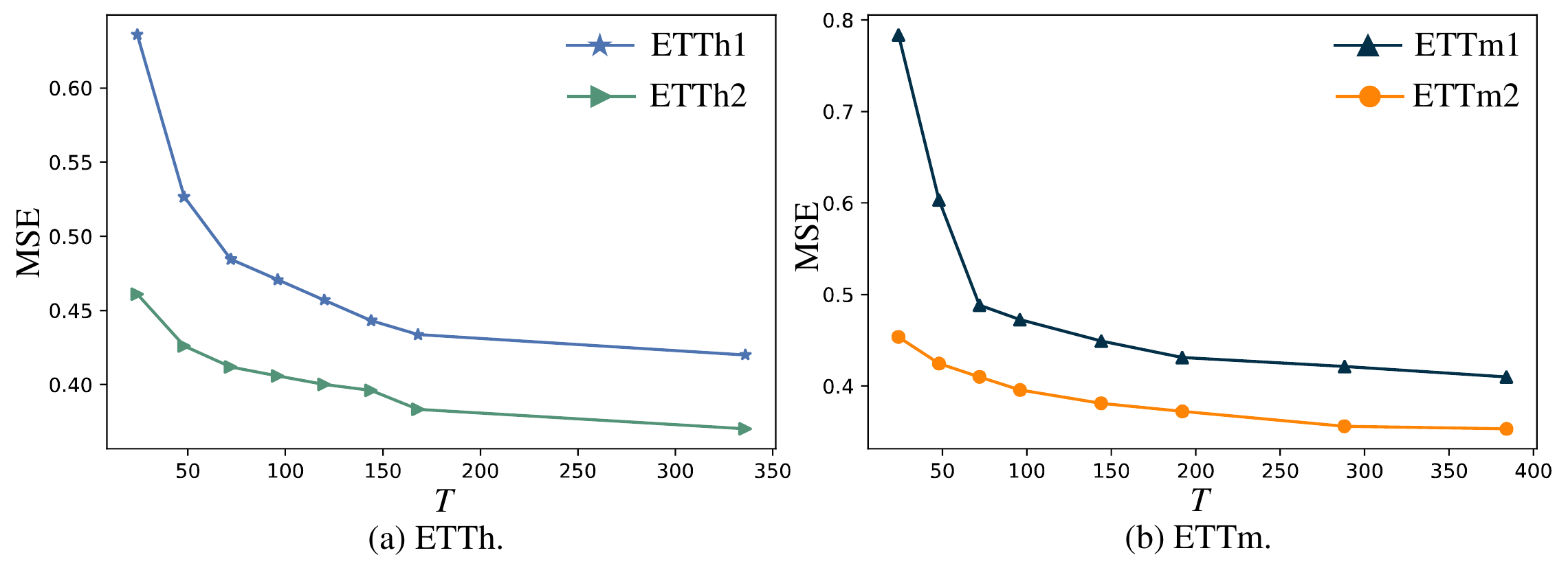}}
	\caption{Results of D-PAD with different lookback window sizes  of long-term forecasting ($H=720$) on \textit{ETT} datasets.}
	\label{fig:input_len}
\end{center}
\end{figure}

{\bf{Lookback window: }}
The size of the lookback window determines how much a model can learn from historical data~\cite{zeng2023transformers}. To study the impact of the lookback window size, we record the results of D-PAD for multivariate long-term forecasting ($H=720$) on hourly granularity datasets (\textit{ETTh1} and \textit{ETTh2}) with $T\in\{24,48,72,96,120,144,168,336\}$ that mean $\{1,2,3,4,5,6,7,14\}$ days. For 15-minute granularity datasets (\textit{ETTm1} and \textit{ETTm2}), we set $T\in\{24,48,72,96,144,192,288,384\}$ that mean $\{6,12,18,24,36,48,72,96\}$ hours. The MSE results are shown in Fig.~\ref{fig:input_len}, from which we can observe that with an increase in input length, the performance of D-PAD shows an upward trend. This suggests that D-PAD can capture more historical information with larger lookback window sizes.

\begin{figure}[tbp]
\begin{center}
	\centerline{\includegraphics[width=\columnwidth]{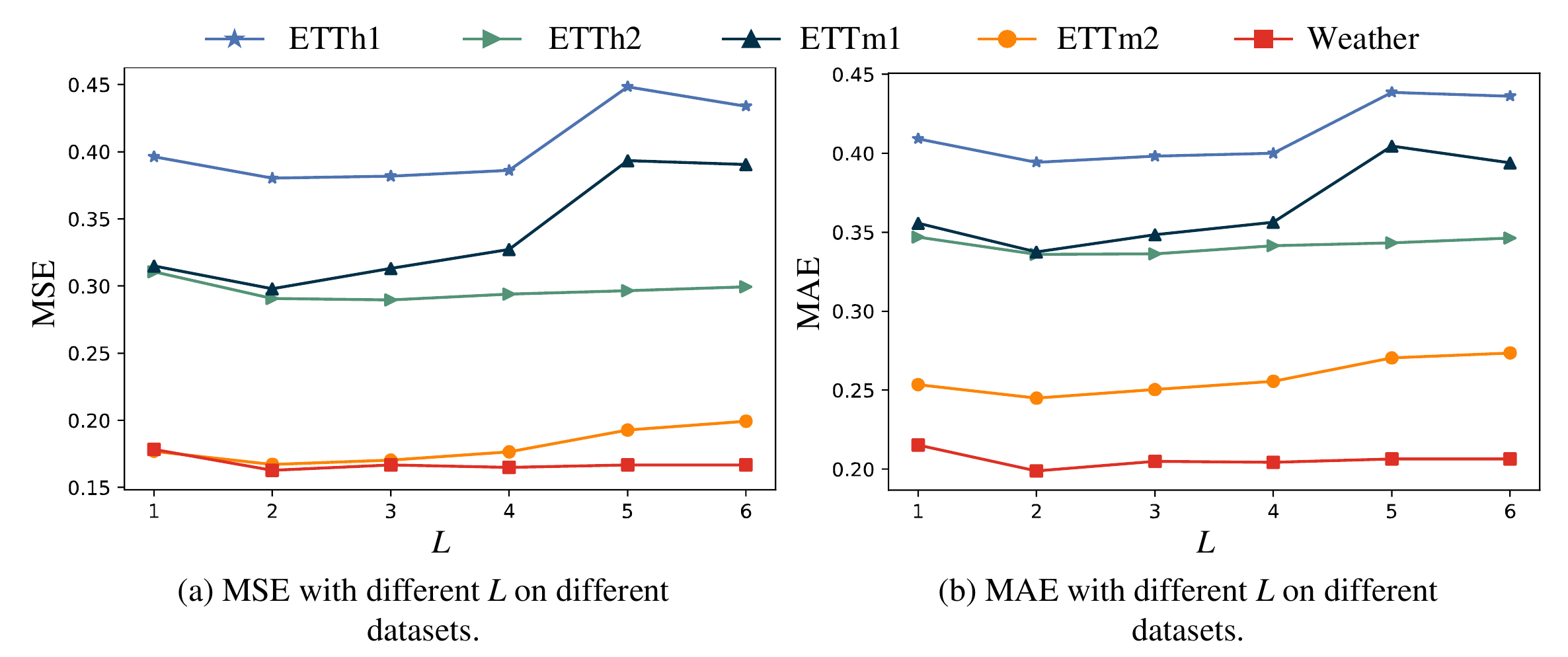}}
	\caption{Results of models with different numbers of levels on \textit{ETT} and \textit{Weather} datasets.}
	\label{fig:level}
\end{center}
\end{figure}

{\bf{D-R-D module levels: }}
With the number of levels $L$ increases in the D-R-D module, the number of components rises exponentially by $2^{L-1}K$, with $K$ being the number of components obtained by an MCD block. To investigate the impact of $L$ on performance, we vary it from 1 to 6 with the step of 1 on \textit{ETT} and \textit{Weather} datasets. The results are shown in Fig.~\ref{fig:level}. The performance of D-PAD first increases and then decreases as $L$ increases, and D-PAD performs the best when $L=2$. This suggests that with multiple components decomposition, a small number of levels can ensure the sufficient disentanglement of time series, while a large number of levels may lead to overfitting.

\subsection{Case Study}
\label{subsec:case}

\begin{figure*}[tbp]
\begin{center}
	\centerline{\includegraphics[width=0.87\linewidth]{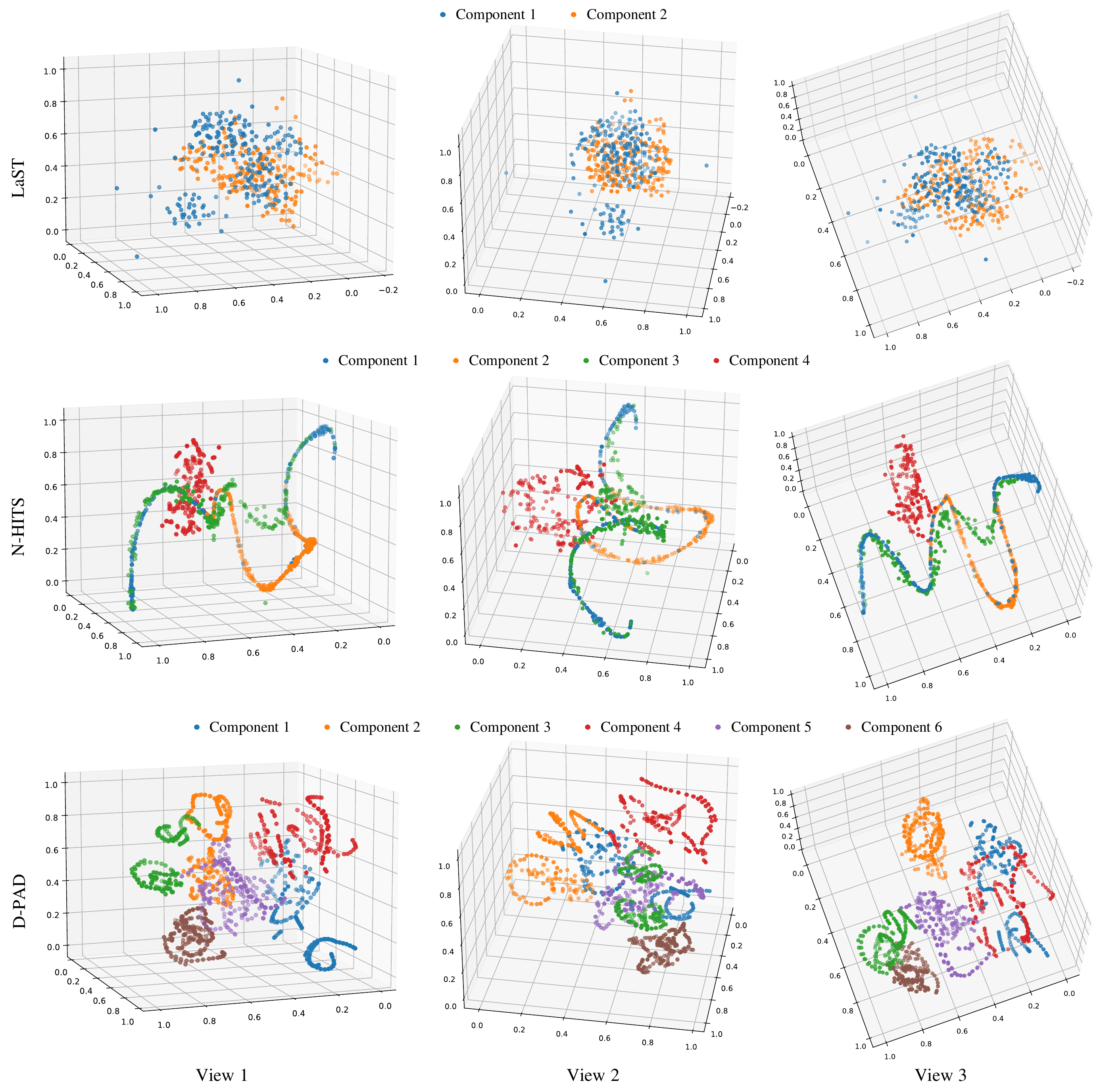}}
	\caption{Visualization of the representations of different components on \textit{ETTh1} dataset from three views.}
	\label{fig:TSNE}
\end{center}
\end{figure*}

{\bf{Representation disentanglement: }}
To exhibit the ability of D-PAD to disentangle intricate temporal patterns, t-SNE~\cite{van2008visualizing} representations of the components obtained by D-PAD and the seasonal-trend and multi-component decomposition SOTA baselines, i.e., LaST~\cite{wang2022learning} and N-HITS~\cite{challu2023nhits}, are plotted in three-dimensional space, as shown in Fig.~\ref{fig:TSNE}. Specifically, by inputting seven consecutive batches on \textit{ETTh1} dataset, components of D-PAD are derived from the output of the D-R-D module, those of LaST from the outputs of its seasonal and trend encoders, and those of N-HITS from the predictions of each stack. We can observe that LaST isolates seasonal and trend components, corresponding to two clusters of points that display minimal spatial distinction. N-HITS presents some mixing in its four-cluster representations, suggesting frequency information dispersion across components. In contrast, D-PAD clearly separates six components with distinct clustering, indicating the effective disentanglement of different frequency patterns.

\begin{figure}[tbp]
\begin{center}
	\centerline{\includegraphics[width=\columnwidth]{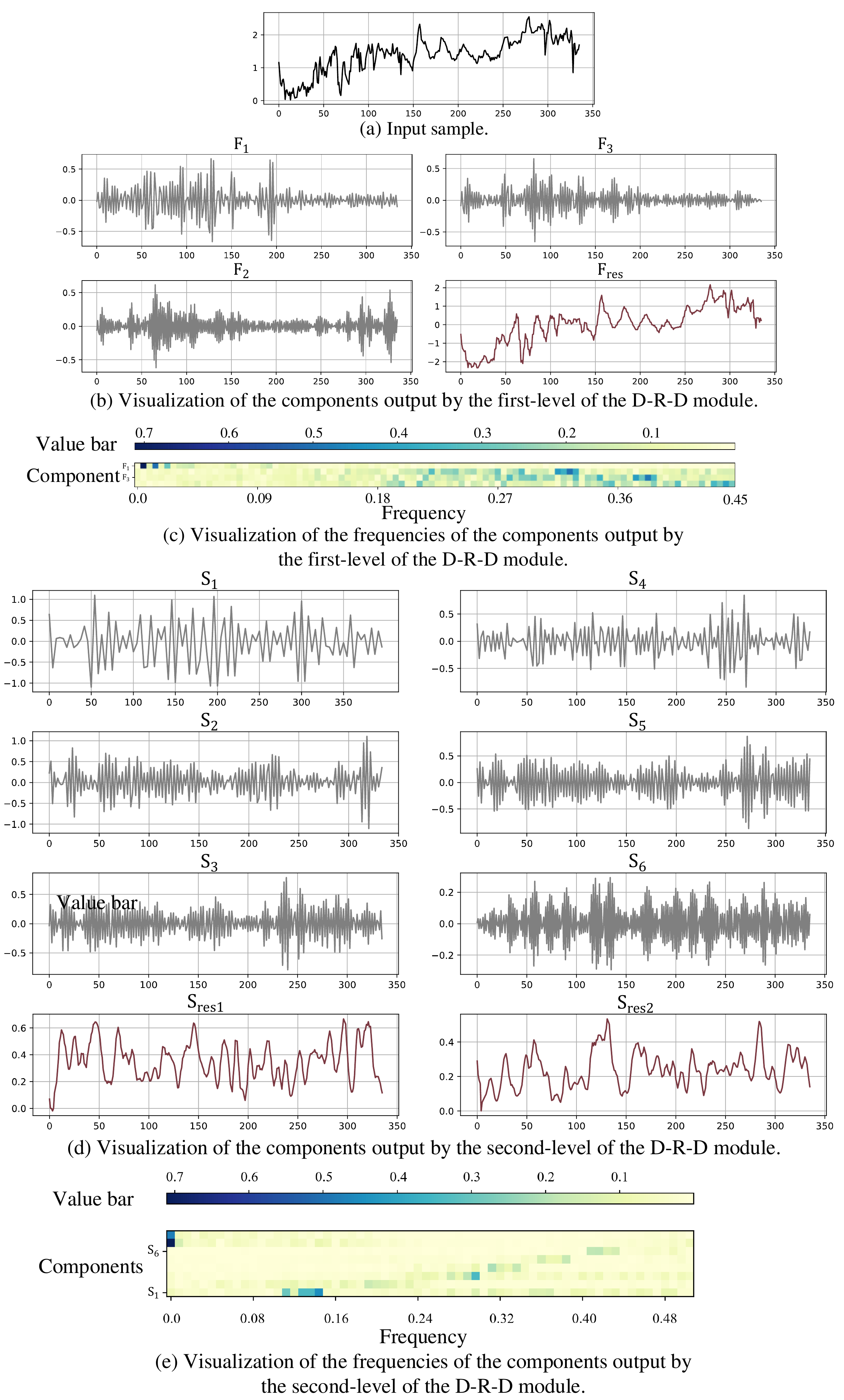}}
	\caption{Visualization of the components and their frequencies output by the D-R-D module on \textit{ETTh1} dataset.}
	\label{fig:midout}
\end{center}
\end{figure}

{\bf{Components analysis: }}
 To intuitively understand the characteristics of the components obtained through the D-R-D module, we visualize the outputs of the first-level and second-level for a sample from \textit{ETTh1} dataset, as shown in Fig.~\ref{fig:midout}(b) and Fig.~\ref{fig:midout}(d), respectively. The input sample is displayed in Fig.~\ref{fig:midout}(a), and the frequency of the outputs of the first and the second level are depicted in Fig.~\ref{fig:midout}(c) and Fig.~\ref{fig:midout}(e), respectively. We can obtain the following observations:

\begin{itemize}
    \item In the first-level output of the D-R-D module, we identify several oscillatory components $\mathrm{F}_{1}$, $\mathrm{F}_{2}$, and $\mathrm{F}_{3}$, alongside a residual component $\mathrm{F}_\mathrm{res}$ capturing the central tendency of the input sequence. These oscillatory components, with near-zero means, reveal diverse dominant frequencies and some mixed patterns. This suggests that while the MCD block aids in data stabilization, the limited separation capability of single-layer decomposition leads to the dispersion of frequency information across components, resulting in the mixing of patterns.
    
    \item The second output of the D-R-D module includes six oscillatory components $\mathrm{S}_{1}$ to $\mathrm{S}_{6}$ and two residuals $\mathrm{S}_\mathrm{res1}$ and $\mathrm{S}_\mathrm{res2}$, each with near-zero means and unique dominant frequencies. This indicates that the D-R-D module can progressively isolate patterns and group similar frequencies together for clearer component distinction.

    \item Apart from the residual components, all components are non-smooth and exhibit obvious oscillations. This indicates that the D-R-D module preserves the original dynamic characteristics of the time series. This is because the interpolation operation used in EMD is discarded, which avoids the introduction of extraneous information.
\end{itemize}

\begin{figure}[tbp]
\begin{center}
	\centerline{\includegraphics[width=\linewidth]{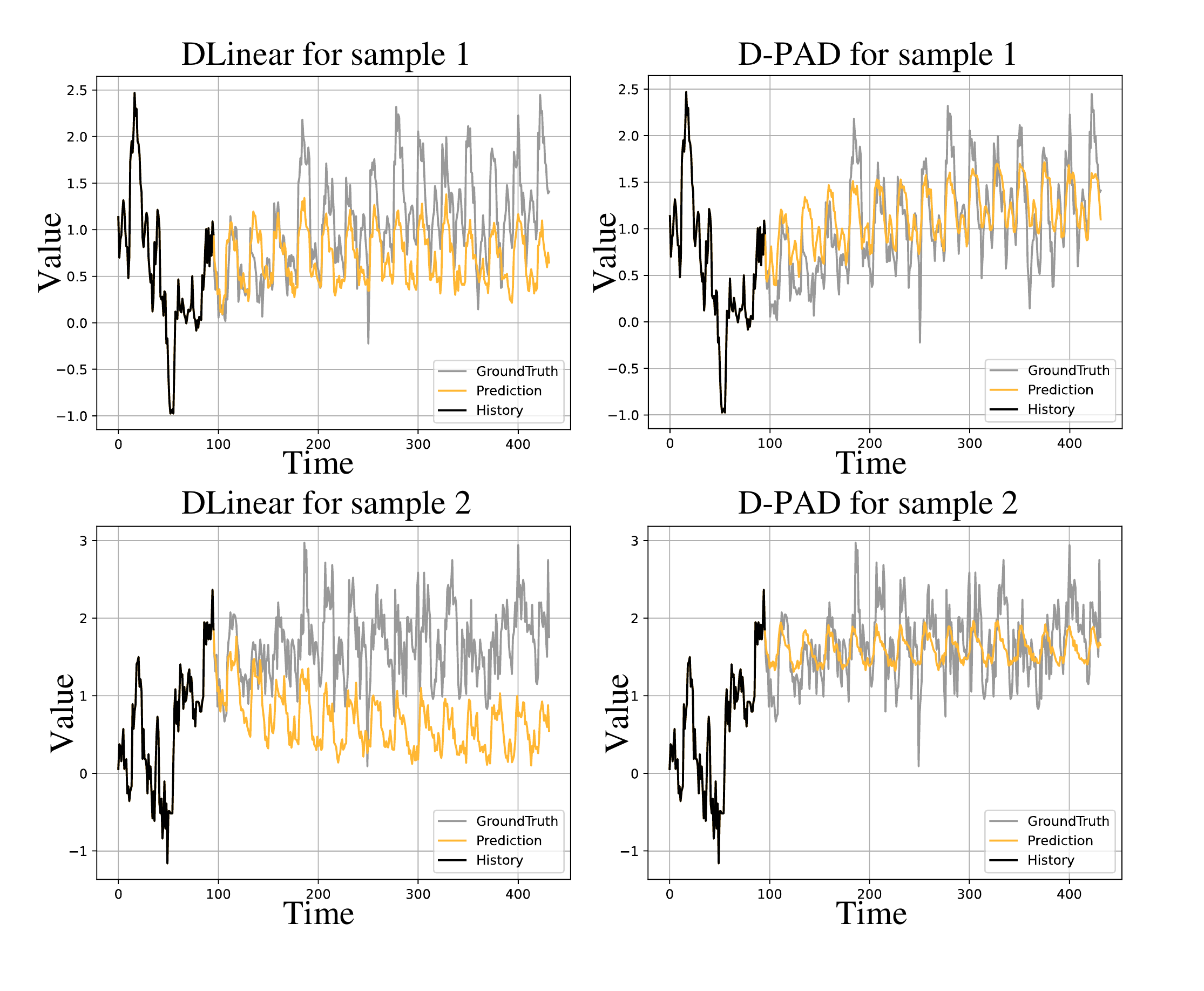}}
	\caption{Visualization of predictions of D-PAD and DLinear on \textit{ETTh1} dataset.}
	\label{fig:MLPLoss-case}
\end{center}
\end{figure}

{\bf{MEMD and stationarity: }}
A classic problem in time series analysis is how to deal with the non-stationarity. Fig.~\ref{fig:MLPLoss-case} shows the predicted curves for two samples exhibiting drifts in the statistical properties, comparing D-PAD with the SOTA decomposition prediction model, i.e., DLinear. For DLinear, there is a noticeable tendency for the predictions to revert towards the mean of historical data. In contrast, D-PAD adapts to the changes in data, providing forecasts that closely align with the actual future values, which indicates that D-PAD can effectively handle non-stationarity in time series. This is because MEMD in the MCD block captures evolving statistical properties through adaptive decomposition, ensuring local characteristics are accurately extracted, and adapts to shifts in statistical properties.

\section{Conclusions and Future Work}
In this paper, we propose D-PAD to disentangle intricate temporal patterns for time series forecasting. Specifically, MCD blocks are introduced to decompose the time series into multiple components with different frequency ranges, and a D-R-D module is proposed to progressively extract the mixed information. The results of extensive experiments show that D-PAD outperforms the SOTA baselines.

In the future, we plan to expand our research in two directions. Firstly, since the morphological operators we use are relatively simple, the ability of model to handle complex or subtle patterns is limited. Future work could explore more intricate SE kernel designs. Secondly, the D-R-D module is pre-set as a binary tree structure, which may limit the flexibility of the model in separating various pattern information, affecting its generalization ability. Future work could explore introducing a more flexible structure for diverse time series.

\bibliographystyle{IEEEtran}
\bibliography{ref}

\newpage

\end{document}